%% file: main.tex
\documentclass[10pt,twocolumn,letterpaper]{article}

\usepackage[pagenumbers]{cvpr} 

\input{preamble}

\definecolor{cvprblue}{rgb}{0.21,0.49,0.74}
\usepackage[pagebackref,breaklinks,colorlinks,allcolors=cvprblue]{hyperref}

\def\paperID{34} 
\def\confName{CVPR}
\def\confYear{2026}

\title{What's Under the Skin? Estimating Swine Body Condition}

\makeatletter
\newcommand{\medium}{\@setfontsize\medium{10.5}{12.5}}
\makeatother

\author*[1]{\fnm{Mk} \sur{Bashar}}\email{basharmk@msu.edu}
\author[2]{\fnm{Kuljit} \sur{Bhatti}}\email{kbhatti2@nebraska.edu}
\author[3]{\fnm{Gary} \sur{Rohrer}}\email{gary.rohrer@usda.gov}
\author[4]{\fnm{Madonna} \sur{Benjamin}}\email{gemus@msu.edu}
\author[2]{\fnm{Tami} \sur{Brown-Brandl}}\email{tami.brownbrandl@unl.edu}
\author[5]{\fnm{Daniel} \sur{Morris}}\email{dmorris@msu.edu}

\affil*[1]{\orgdiv{Computer Science and Engineering}, \orgname{Michigan State University}, \orgaddress{\city{East Lansing}, \state{MI}, \country{USA}}}
\affil[2]{\orgdiv{Biological System Engineering}, \orgname{University of Nebraska-Lincoln}, 
\orgaddress{\city{Lincoln}, \state{NE}, \country{USA}}}
\affil[3]{\orgdiv{USDA ARS}, \orgname{Clay Center Nebraska}, \orgaddress{\city{Clay Center}, \state{NE}, \country{USA}}}
\affil[4]{\orgdiv{Large Animal Clinical Sciences}, \orgname{Michigan State University}, \orgaddress{\city{East Lansing}, \state{MI}, \country{USA}}}
\affil[5]{\orgdiv{Biosystems and Agricultural Engineering}, \orgname{Michigan State University}, \orgaddress{\city{East Lansing}, \state{MI}, \country{USA}}}

\makeatletter
\gdef\@snemails{\{basharmk, gemus, dmorris\}@msu.edu, gary.rohrer@usda.gov, kbhatti2@nebraska.edu, tami.brownbrandl@unl.edu}
\makeatother
\begin{document}

\maketitle
\input{sec/0_body}    
{
    \small
    \bibliographystyle{ieeenat_fullname}
    \bibliography{main}
}

\input{sec/X_suppl}

\end{document}

%% file: preamble.tex
\usepackage{comment}
\newif\ifblinded
\blindedfalse

\ifblinded
  \excludecomment{unblind}
  \includecomment{blind}
\else
  \includecomment{unblind}
  \excludecomment{blind}
\fi

\newcommand{\blindswitch}[2]{%
  \ifblinded
    \colorbox{yellow}{[\,#1\,]}%
  \else
    #2%
  \fi
}

\makeatletter
\providecommand{\fnm}[1]{#1}
\providecommand{\sur}[1]{#1}

\providecommand{\equalcont}[1]{}
\providecommand{\orgdiv}[1]{#1}
\providecommand{\orgname}[1]{#1}
\providecommand{\orgaddress}[1]{#1}

\providecommand{\city}[1]{#1}
\providecommand{\state}[1]{#1}

\providecommand{\country}[1]{#1}

\def\@snauthors{}
\def\@snaffils{}
\def\@snemails{}
\newif\if@snfirstau \@snfirstautrue
\newif\if@snfirstaf \@snfirstaftrue
\newif\if@snfirstem \@snfirstemtrue

\let\@cvprauthor\author
\renewcommand{\author}{\@ifstar{\@snaddauth}{\@snaddauth}}
\newcommand{\@snaddauth}[2][1]{%
  \if@snfirstau
    \gdef\@snauthors{#2$^{#1}$}\@snfirstaufalse
  \else
    \xdef\@snauthors{\@snauthors, #2$^{#1}$}%
  \fi
}

\providecommand{\email}[1]{}
\renewcommand{\email}[1]{%
  \if@snfirstem
    \gdef\@snemails{#1}\@snfirstemfalse
  \else
    \xdef\@snemails{\@snemails, #1}%
  \fi
}

\newcommand{\affil}{\@ifstar{\@snaddaffil}{\@snaddaffil}}
\newcommand{\@snaddaffil}[2][1]{%
  \if@snfirstaf
    \gdef\@snaffils{$^{#1}$#2}\@snfirstaffalse
  \else
    \xdef\@snaffils{\@snaffils\noexpand\\ $^{#1}$#2}%
  \fi
}

\let\@cvprmaketitle\maketitle
\renewcommand{\maketitle}{%
  \@cvprauthor{%
    \parbox{\textwidth}{\centering
      \@snauthors\\[2pt]
      \@snaffils\\[2pt]
      {\tt\small \@snemails}%
    }%
  }%
  \@cvprmaketitle
}
\makeatother

%% file: sec/0_body.tex
\begin{abstract}
Sow body condition is an important indicator for growers as it has a large impact on lactation performance and piglet survival.  However, body condition measures used during production, such as visual scoring and calipers, correlate poorly with underlying tissue composition.  Ultrasound scans can provide direct measurements of subcutaneous backfat thickness and loin muscle depth, but their operation is labor intensive and not scalable for production.  We present PigFormer, an end-to-end two-stage system that takes raw depth frames from a ceiling-mounted RGB-D camera and predicts subcutaneous backfat thickness, loin muscle depth, and total tissue thickness at the last rib. Stage 1 is a geometric front-end that converts raw depth into a standardized height map via SAM3-to-MaskDINO segmentation distillation, ground-plane removal, and orientation normalization. Stage 2 is a Slice Attention Encoder that treats each height map as a sequence of cross-sectional slices and captures spatial relationships along the full dorsal surface. On a multi-site dataset of 319 sow and gilt instances from two facilities, PigFormer achieves 2.43\,mm backfat MAE and 3.87\,mm overall MAE. It outperforms strong single-stage ResNet-18 and ViT-small baselines. PigFormer offers a practical path toward continuous, automated, non-contact body condition monitoring in commercial swine production. Code is available at \url{https://github.com/iambashar/Pigformer}.
\end{abstract}

\section{Introduction}
\label{sec:intro}
In sow and gilt management, body condition is linked to nutritional supply and to the number of healthy piglets that survive to weaning. Adequate backfat provides energy reserves for lactation~\cite{Nuntapaitoon2024Backfat,Gourley2020PigletSurvival}, and sufficient loin depth reflects muscle development that supports both the dam and her offspring~\cite{Authement2023LoinSurvival}. Producers need tools that reveal whether animals are becoming too fat or too thin and that support timely feeding decisions before inefficiency accumulates. Feed is the dominant variable cost in pork production, so modest improvements in feed allocation affect farm-level economics~\cite{PIGFeedCost}.

Current methods for performing body condition scoring (BCS) are imprecise. Visual appraisal correlates poorly with actual tissue composition (r$^{2} = 0.19$ for visual scores vs.\ backfat across 731 sows~\cite{Young2001BCSBackfat}), while indirect tools such as the sow caliper suffer from low inter-observer repeatability and limited agreement with ultrasound references~\cite{Knauer2007Caliper,knauer2015sow,Li2021SowCaliperEval}. A more direct measure of body condition is provided by ultrasound-derived continuous traits: subcutaneous backfat thickness, loin muscle depth, and their sum (total tissue thickness). These quantities capture the fat and muscle compartments that predict lactation performance and piglet survival~\cite{Fernandes2020BodyComposition,Nuntapaitoon2024Backfat,Authement2023LoinSurvival}. While ultrasound is the accepted reference standard for these traits, routine collection at production scale requires skilled operators, specialized equipment, and animal handling that limit measurement frequency~\cite{Jian2024BackfatMachineVision}. A gap remains between the measurements that best guide feed management and those that farms can collect at scale.

This gap motivates a low-cost, automatic, non-contact alternative. Computer vision can capture repeated phenotypes in livestock settings with less handling stress~\cite{Fernandes2019Automated}. Prior RGB-D and 3D systems in pigs have focused on body-weight and growth monitoring~\cite{Fernandes2019Automated,Yu2021Forecasting}. Recent work has begun estimating sow backfat thickness or body condition from images~\cite{Jian2024BackfatMachineVision,Xue2023CATCBAM}. End-to-end RGB-D estimation of both backfat and loin-related targets for practical farm monitoring remains limited.

We study multi-target regression from RGB-D observations of standing sows and gilts, focusing on the measurements producers use in feed decisions: backfat, loin, and total tissue thickness. We introduce PigFormer, a two-stage system that takes raw depth frames as input: Stage 1 is a geometric front-end that converts raw RGB-D recordings into normalized height maps preserving dorsal geometry while removing appearance variation unrelated to body composition; Stage 2 is a Slice Attention Encoder that treats each height map as a sequence of cross-sectional slices and regresses all three targets jointly.

Our contributions are threefold. First, we are the first to formulate automated sow body-condition estimation from RGB-D data as a fully automated, end-to-end trainable multi-target regression problem built around production-relevant measurements. Second, we propose PigFormer, an end-to-end two-stage system. Stage 1 (geometric front-end) uses SAM3-to-MaskDINO segmentation distillation, ground-plane removal, and orientation normalization to standardize noisy field recordings into a clean geometric representation. Stage 2 (Slice Attention Encoder) uses rotary position embeddings and dual pooling to regress all three body-condition targets jointly. Third, on a multi-site dataset of 319 instances PigFormer achieves an overall MAE of 3.87\,mm, outperforming strong single-stage ResNet-18 and ViT-small baselines by 22\% and 39\% respectively. Stage 1's geometric inductive bias therefore carries useful information on top of pretrained backbone features. \Cref{sec:method} details PigFormer's two stages, \Cref{sec:experiments} reports empirical results, and \Cref{sec:discussion} discusses implications for on-farm practice.

\section{Related Work}
\label{sec:related}

\paragraph{Swine imaging and body condition.}
Swine vision work can be grouped by target trait. For body weight and growth, RGB-D and structured-light systems use 3D point clouds or depth maps with multi-output CNNs or transformer-style regressors to estimate live weight and linear body dimensions in both experimental and on-farm settings~\cite{Fernandes2019Automated,Yu2021Forecasting,Zhang2021MultiOutput,He2023TwoStream}. Extending beyond gross weight, Fernandes \etal~\cite{Fernandes2020BodyComposition} showed that top-view 3D images of finishing pigs can predict carcass muscle depth and backfat, with deep models outperforming conventional analytic pipelines. For sow body condition, image-based approaches have focused mainly on continuous backfat thickness (in mm) or discrete visual body condition scores. Yu \etal and Li \etal~\cite{Yu2022NonContactBackfat,Li2023CNNViTBackfat} regress pregnant sows’ ultrasonic backfat thickness from single 2D back-view images using CNN and CNN–ViT hybrids. Depth-based methods exploit hindquarter morphology: Teng \etal~\cite{Teng2018KinectSow} use Kinect-derived 3D buttock shape to predict categorical sow body condition scores on a five-grade visual scale, while Jian \etal~\cite{Jian2024BackfatMachineVision} build a machine-vision system that regresses sow backfat thickness in millimeters from buttock morphology. CNN–transformer hybrids such as CAT-CBAM-Net predict five-category sow body condition scores (too lean, slightly lean, ideal, slightly fat, too fat) from back-view RGB images with high classification accuracy~\cite{Xue2023CATCBAM}. Closest to our formulation, Peppmeier \etal~\cite{Peppmeier2023Ultrasound} simultaneously estimate backfat depth, loin eye depth, and intramuscular fat percentage from longitudinal ultrasound images, providing a multi-target precedent but still requiring specialized ultrasound hardware and restraint. To the best of our knowledge, we are the first to propose a fully automated, end-to-end trainable pipeline utilizing a transformer architecture to directly estimate multiple sow body-condition targets—backfat, loin, and total tissue thickness—from RGB-D recordings.

\paragraph{Cattle body condition scoring.} In dairy cattle, automated BCS has received more attention and provides useful methodological lessons. Earlier systems relied on handcrafted geometric cues from depth images, such as anatomical landmarks and surface curvature~\cite{Zhao2020DairyBCS,RodriguezAlvarez2019DairyBCS}. Subsequent work moved toward deeper models and field validation, including commercial-system evaluation showing strong correlation with manual scoring but weaker performance at score extremes~\cite{Mullins2019CommercialBCS}. More recent work has broadened the setting. Depth sensing has been applied to body weight and coarse BCS in beef cows~\cite{Xiong2023BeefBCS}, and head-to-head comparisons suggest point clouds do not outperform depth images for dairy BCS prediction~\cite{Tang2026PointCloudBCS}. These cattle studies inform our design choices around representation and evaluation, but they target categorical BCS rather than the continuous backfat and loin regression we study here.

\begin{figure*}[t]
  \centering
  \includegraphics[width=\linewidth]{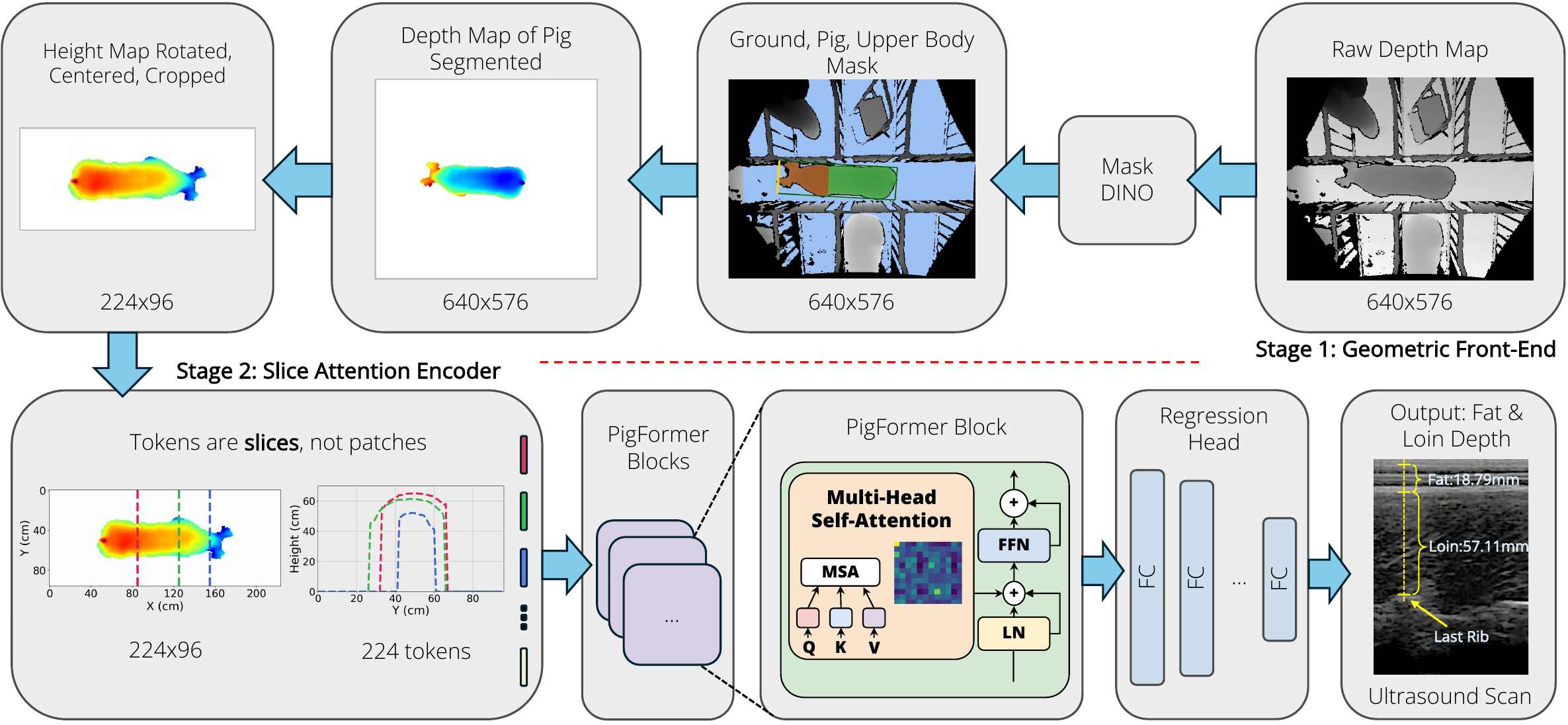}
  \caption{Overview of PigFormer, a two-stage system. A ceiling-mounted RGB-D camera captures raw depth frames. \textbf{Stage 1} (geometric front-end): SAM3-to-MaskDINO distillation segments the pig, upper body, and ground from the depth channel; ground-plane subtraction and binning produce a centered height map oriented with the pig facing right. \textbf{Stage 2} (Slice Attention Encoder): tokenizes the 224 cross-sectional slices of the height map and regresses backfat thickness, loin muscle depth, and total tissue depth at the last rib. The system is trained on single frames with Huber loss on backfat and loin measurements obtained from ultrasound scans of each pig, 
  and averages all-frame predictions at inference.
  }
  \label{fig:pipeline}

\end{figure*}

\section{Method}
\label{sec:method}
\subsection{Data and labels}
We collect RGB-D recordings at two \blindswitch{facilities}{U.S.\ facilities} (\blindswitch{Site 1}{Michigan State University} and \blindswitch{Site 2}{the U.S.\ Meat Animal Research Center in collaboration with the University of Nebraska--Lincoln}) using ceiling-mounted Orbbec depth cameras. Ground-truth backfat thickness and loin muscle depth at the last rib are obtained via B-mode ultrasound by trained operators. This procedure requires expensive equipment, skilled personnel, and physical restraint, which limits the scale of labeled data collection. Some animals were scanned at multiple dates spanning 4--6 months; because body composition changes over this interval, we treat each scan date as an independent instance. The combined dataset comprises 319 instances (116 \blindswitch{Site 1}{MSU}, 203 \blindswitch{Site 2}{UNL}) totaling 6{,}705 depth frames.

\subsection{PigFormer}
\label{sec:pigformer}
PigFormer is a fully automated, end-to-end two-stage system that takes a raw $576 \times 640$ 16-bit depth frame from a ceiling-mounted camera and predicts subcutaneous backfat thickness, loin muscle depth, and total tissue depth at the last rib. \textbf{Stage 1} is a geometric front-end that segments the pig from its surroundings and normalizes camera pose and animal heading to produce a standardized height map. \textbf{Stage 2} is a Slice Attention Encoder that processes the height map as a sequence of cross-sectional slices and jointly regresses all three targets.

\subsubsection{Stage 1: Geometric Front-End}
Stage 1 converts each raw depth frame into a standardized $96 \times 224$ height map in four steps. (1)~SAM3-to-MaskDINO segmentation distillation~\cite{Carion2025SAM3,li2023maskdino} predicts whole-pig, upper-body, and ground masks from the depth channel alone. (2)~We estimate the ground plane and subtract it to obtain height above ground. (3)~We project the pig points onto the ground plane and bin them at $1\,\text{cm} \times 1\,\text{cm}$ resolution. (4)~Orientation normalization, via minimum-area bounding rectangle fitting and upper-body heading, rotates each map so the pig faces to the right. The resulting height map eliminates confounding variation from camera pose, lighting, and visual markings, letting Stage 2 focus on dorsal shape. Full Stage-1 details and dataset statistics are provided in the supplementary material (\cref{sec:supp_preprocess}); two faster Stage 1 alternatives (a pruned MaskDINO and a UNet) are described in \cref{sec:supp_pruned_maskdino} and evaluated in \cref{sec:experiments}.

\subsubsection{Stage 2: Slice Attention Encoder}
Dorsal body-condition targets, such as backfat and loin depth, depend on shape relationships that span the full back surface. The Slice Attention Encoder (Stage 2) captures this long-range spatial structure using rotary position embeddings applied to the height map produced by Stage 1.

\textbf{Sequence representation.} Stage 2 treats the $96 \times 224$ height map as a sequence of 224 tokens. Each token encodes one column of the height map from left to right. Column tokens outperformed patch-based tokens (typical for images) in our experiments.   

\textbf{Rotary position embedding.} Rather than adding learned or sinusoidal positional encodings to the input, Stage 2 applies Rotary Position Embedding (RoPE)~\cite{Su2021RoPE} to the query and key vectors in the attention mechanism. Each position index generates a rotation matrix from sinusoidal frequencies; query and key vectors are split into dimension pairs, and each pair is rotated by an angle proportional to the position. The resulting dot product depends on the relative distance between two cross-sections, giving the model a smooth sense of spatial proximity along the spine. Rib-site measurements depend on knowing \emph{where} along the body the model attends, making relative position awareness critical.

\textbf{Encoder.} Each encoder layer consists of multi-head self-attention with RoPE applied to queries and keys, followed by a feed-forward network (Linear $\rightarrow$ GELU $\rightarrow$ Linear) with layer normalization and dropout. Self-attention allows every cross-sectional slice to attend to every other slice; for example, the rib region can contextualize its features by looking at the loin region. After encoding, each token's representation captures its local height profile and its relationship to the rest of the body.

\textbf{Dual pooling.} After the encoder, the 224 token representations are collapsed into a fixed-size vector per animal via mean and max pooling along the sequence dimension, yielding a 192-dimensional summary. Mean pooling captures overall body condition, while max pooling preserves prominent anatomical features. Concatenating both yields a richer summary than either pooling strategy alone.

\textbf{Prediction head.} The pooled vector passes through a layer-normalized MLP head (Linear $\rightarrow$ GELU $\rightarrow$ Dropout $\rightarrow$ Linear) that outputs three regression targets: backfat depth, loin muscle depth, and total tissue depth at the last rib, all in millimeters.

\textbf{Loss function.} We train Stage 2 with a Huber loss applied independently to each of the three targets, with $y_t = y_f + y_l$. Supervising the sum as an explicit third target adds a consistency signal between fat and loin predictions:
{\setlength{\abovedisplayskip}{2pt}\setlength{\belowdisplayskip}{2pt}
\begin{equation}
  \mathcal{L} = \mathcal{L}_{\text{Huber}}(y_f, \hat{y}_f) + \mathcal{L}_{\text{Huber}}(y_l, \hat{y}_l) + \mathcal{L}_{\text{Huber}}(y_t, \hat{y}_t),
  \label{eq:loss}
\end{equation}
\begin{equation}
  \mathcal{L}_{\text{Huber}}(y, \hat{y}) =
  \begin{cases}
    \tfrac{1}{2}(y - \hat{y})^{2}, & |y - \hat{y}| \leq \delta, \\
    \delta\,|y - \hat{y}| - \tfrac{1}{2}\delta^{2}, & \text{otherwise}.
  \end{cases}
  \label{eq:huber}
\end{equation}}%
Huber is less sensitive to noise in ultrasound-based measurements than $L_2$ and penalizes large errors more than $L_1$; we use $\delta = 1.0$\,mm.

\textbf{Training and inference.} Each epoch we sample a single random frame per animal; over thousands of epochs Stage 2 sees diverse poses without the blurring of pixel-level averaging. At inference, we forward every frame independently and average the predictions.

\subsection{Single-stage baselines}
\label{sec:baselines}
To quantify how much PigFormer's geometric front-end contributes on top of a strong pretrained backbone, we compare PigFormer against two single-stage systems. Both skip Stage 1 and feed raw $576 \times 640$ 16-bit depth frames directly to an ImageNet-pretrained backbone: \textbf{ResNet-18} at native camera resolution, and \textbf{ViT-small} (\texttt{vit\_small\_patch16\_224}) bilinear-resized to $224 \times 224$ inside the forward pass (its fixed-grid positional embeddings preclude native-resolution input). Both backbones receive a 1-channel \texttt{conv1}/patch-embedding surgery that averages the pretrained kernel across the input channel axis. The only adjustment to the raw depth is $z$-score normalization with training-fold statistics---no segmentation, no ground-plane removal, no orientation normalization, no crop. This is the minimum unavoidable scale adjustment: the ImageNet-pretrained backbones expect inputs in a specific regime. Statistics are computed once on the training fold and frozen for validation and test, eliminating distributional leakage. The regression head predicts fat and loin only ($\hat{y}_t = \hat{y}_f + \hat{y}_l$ at evaluation). For a fair frame-set comparison, both training and evaluation use the same per-bag frame indices PigFormer uses, recovered from Stage 1's output. These baselines isolate the contribution of Stage 1 from the choice of Stage-2 architecture and pretrained features. Additional Stage-2 architecture ablations on the height map (MLP and CNN encoders, both consuming Stage 1's output) are reported in \cref{sec:supp_arch}.

\vspace{-0.5em}
\section{Experiments}
\label{sec:experiments}
\subsection{Evaluation protocol}
We evaluate performance with mean absolute error (MAE) for each anatomical target: backfat depth, loin muscle depth, and total tissue depth at the last rib. For model selection and early stopping based on overall MAE, we use a validation subset partitioned from the training instances. A separate test set is used strictly for reporting final results. Because multiple frames can be acquired from the same animal, we enforce identity-level separation across all splits: validation and test animals do not appear in the training set. This protocol reflects the practical deployment setting, where the goal is to generalize to unseen animals rather than memorize repeated views from the same individual.

\begin{table*}[t]
  \caption{Held-out test results on 79 sow / gilt instances. MAE in mm; MAPE $=$ MAE\,/\,mean. Per-frame inference is measured on A100 with batch $=$ 1 (MaskDINO Stage 1 in fp16; UNet Stage 1, the single-stage backbones, and PigFormer Stage 2 in fp32). Single-stage ViT-small and ResNet-18 baselines feed raw $576 \times 640$ depth directly to an ImageNet-pretrained backbone with only $z$-score normalization and predict fat and loin only ($\hat{y}_t = \hat{y}_f + \hat{y}_l$ at evaluation). We report three Stage 1 variants for PigFormer: the original MaskDINO (R50-300q-9L), a pruned MaskDINO (R18-50q-5L), and a UNet (MobileNetV3-Small encoder); the pruned and UNet variants retrain Stage 2 on their height maps. PigFormer numbers are 4-fold cross-validation ensembles with output aggregation. Best MAE and MAPE in \textbf{bold}.}
  \label{tab:results}
  \centering
  \resizebox{\textwidth}{!}{
  \begin{tabular}{@{}l cc cccc cccc@{}}
    \toprule
    & \multicolumn{2}{c}{Inference (ms\,/\,frame)} & \multicolumn{4}{c}{MAE (mm) $\downarrow$} & \multicolumn{4}{c}{MAPE $\downarrow$} \\
    \cmidrule(lr){2-3} \cmidrule(lr){4-7} \cmidrule(lr){8-11}
    Method & Stage 1 & Stage 2 & Fat & Loin & Total & Overall & Fat & Loin & Total & Overall \\
    \midrule
    ViT-small (single-stage) & --- & 4.98 & 3.57 & 7.29 & 8.16 & 6.34 & 21.4\% & 13.6\% & 11.5\% & 13.5\% \\
    ResNet-18 (single-stage) & --- & 2.88 & 2.88 & 6.10 & 5.81 & 4.93 & 17.3\% & 11.4\% & 8.2\% & 10.5\% \\
    \midrule
    \textbf{PigFormer} w. MaskDINO & 106.92 & 0.50 & 2.43 & \textbf{5.01} & \textbf{4.19} & \textbf{3.87} & 14.6\% & \textbf{9.3\%} & \textbf{5.9\%} & \textbf{8.2\%} \\
    \textbf{PigFormer} w. Pruned MaskDINO & 52.73 & 0.50 & \textbf{2.34} & 5.27 & 4.20 & 3.94 & \textbf{14.1\%} & 9.8\% & 5.9\% & 8.4\% \\
    \textbf{PigFormer} w. UNet & 6.58 & 0.50 & 2.40 & 5.20 & 4.26 & 3.95 & 14.4\% & 9.7\% & 6.0\% & 8.4\% \\
    \midrule
    Human Ultrasound Std & --- & --- & 1.30 & 2.02 & 2.29 & 1.87 & --- & --- & --- & --- \\
    \bottomrule
  \end{tabular}
  }
\end{table*}

\subsection{Results}
\Cref{tab:results} compares PigFormer against the two single-stage baselines, reporting both mean absolute error (MAE, mm) and mean absolute percentage error (MAPE = MAE\,/\,mean). PigFormer (with the original MaskDINO Stage 1) achieves an overall MAE of 3.87\,mm (4-fold ensemble, output aggregation), reducing error by 22\% relative to single-stage ResNet-18 (4.93\,mm) and by 39\% relative to single-stage ViT-small (6.34\,mm). Its fat MAE of 2.43\,mm is within twice the variability of repeated manual ultrasound (1.30\,mm).

The two single-stage baselines isolate the contribution of PigFormer's Stage 1. The stronger of them---ResNet-18 fed at native $576 \times 640$ resolution---lags the full two-stage PigFormer by 1.06\,mm overall (4.93 vs 3.87\,mm), with most of the gap concentrated on loin (6.10 vs 5.01\,mm) rather than fat (2.88 vs 2.43\,mm). This pattern matches the role of Stage 1's orientation normalization and ground-plane removal. Backfat depends on a localized rib-region cue that pretrained features can recover from raw depth. Loin estimation benefits more from explicit heading alignment that places the rib site at a canonical location. Single-stage ViT-small lags further (6.34\,mm) for two reasons. First, its fixed-grid positional embeddings force a destructive bilinear resize from $576 \times 640$ to $224 \times 224$, discarding spatial detail that ResNet-18 retains via \texttt{AdaptiveAvgPool2d}. Second, at 240 training instances, the ViT's weaker locality bias overfits earlier than the ResNet's.

\textbf{Stage 1 alternatives and the speed--accuracy trade-off.} The original Stage 1 (MaskDINO with R50 backbone, 300 object queries, 9 decoder layers) dominates PigFormer's wall-clock cost: Stage 2 itself takes 0.50\,ms / frame, so Stage 1 dominates the latency budget. We chose a detection-based segmenter because production deployments encounter non-pig content (handlers, feeders, partial pigs at the FOV edge, empty pens). A detection-style head trained to suppress ``no-object'' matches has stronger out-of-distribution behavior than a foreground-only segmenter. To check whether the full-size segmenter is necessary at the achieved accuracy, we evaluate two faster Stage 1 alternatives in \cref{tab:results}, each paired with a Stage 2 retrained on its respective height maps. The first is a \emph{pruned} MaskDINO that preserves the detection inductive bias by cutting the backbone (R50$\rightarrow$R18), the object queries ($300 \rightarrow 50$, sufficient for one-pig-per-frame), and the decoder depth ($9 \rightarrow 5$ layers). The second is a UNet that drops the detection machinery for additional speed. Both reach the same overall MAE as the original within $+0.08$\,mm (3.94 and 3.95 vs.\ 3.87\,mm) while cutting Stage 1 wall-clock time by $2\times$ and $16\times$. PigFormer with the UNet Stage 1 runs end-to-end in $\approx 7$\,ms / frame on A100, fast enough to track a single-camera real-time depth stream. \Cref{sec:supp_pruned_maskdino} details the pruning and architecture.

Full ablations on six design axes, a classical-ML lower bound that rules out global-statistic regression, and an input-gradient analysis localizing the encoder's attention to the tail/rump and rib-shoulder regions are in \cref{sec:supp_ablation,sec:supp_classical,sec:supp_attention}.

\vspace{-.5em}
\section{Limitations}
\label{sec:limitations}
The dataset is limited to 319 instances, constrained by the cost of ultrasound acquisition, which requires trained personnel, specialized equipment, and animal restraint. Although cross-site diversity proved sufficient for effective training, this scale limits statistical power and may reduce generalization to breeds or management regimes not represented in either site. No public RGB-D sow body-condition benchmark with matched rib-site ultrasound labels exists. This prevents direct external comparison, so we evaluate against raw-depth baselines on the same animals.

\textbf{True cross-site generalization.} \Cref{tab:abl_site} shows that single-site training generalizes poorly to the mixed test pool. A stronger generalization probe---training on one site and evaluating only on the other site's test portion---would isolate transfer across breed and management regime more directly. With only two sites and 26 / 53 test instances per site, the resulting estimates would be dominated by sampling noise on top of a known distribution shift (\cref{tab:dataset_stats}); the credible test is recording at a third facility. Future work will pursue a third-site collection so the cross-site claim can be evaluated against held-out farms rather than held-out portions of the existing two.

\vspace{-0.5em}
\section{Discussion and Conclusion}
\label{sec:discussion}
PigFormer offers a non-contact alternative to manual ultrasound that runs on a ceiling-mounted depth camera. Its 3.87\,mm overall MAE sits within twice the intra-operator variability of repeated manual ultrasound (1.30 / 2.02 / 2.29\,mm for fat / loin / total), and its 2.43\,mm fat error approaches the noise floor of the reference standard. With the UNet Stage 1 alternative, end-to-end inference takes $\approx 7$\,ms / frame on a single A100 GPU, fast enough for real-time monitoring on an installed-camera stream. Automated RGB-D body-condition screening therefore reaches an accuracy and latency regime usable for routine on-farm deployment, with the choice of Stage 1 variant trading off out-of-distribution robustness against speed depending on the deployment context. We will release the training and evaluation code together with the ultrasound annotation tool.

\section{Acknowledgements}
\blindswitch{Acknowledgements to be added if accepted.}{We thank the MSU Data Machine, which is supported through the NSF Campus CyberInfrastructure program through grant \#2200792. We also thank the USMARC and MSU experts for their collaboration on data collection and annotation. This work was supported by the USDA Swine Phenotyping Grant: USDA NIFA award \#2022-67021-37858}. USDA is an equal opportunity employer.

%% file: sec/X_suppl.tex
\clearpage
\setcounter{page}{1}
\maketitlesupplementary

\section{Stage 1: Geometric Front-End Details}
\label{sec:supp_preprocess}

This section expands on Stage 1 of PigFormer summarized in \cref{sec:pigformer} of the main paper.

\begin{figure}[t]
  \centering
  \includegraphics[width=0.47\textwidth]{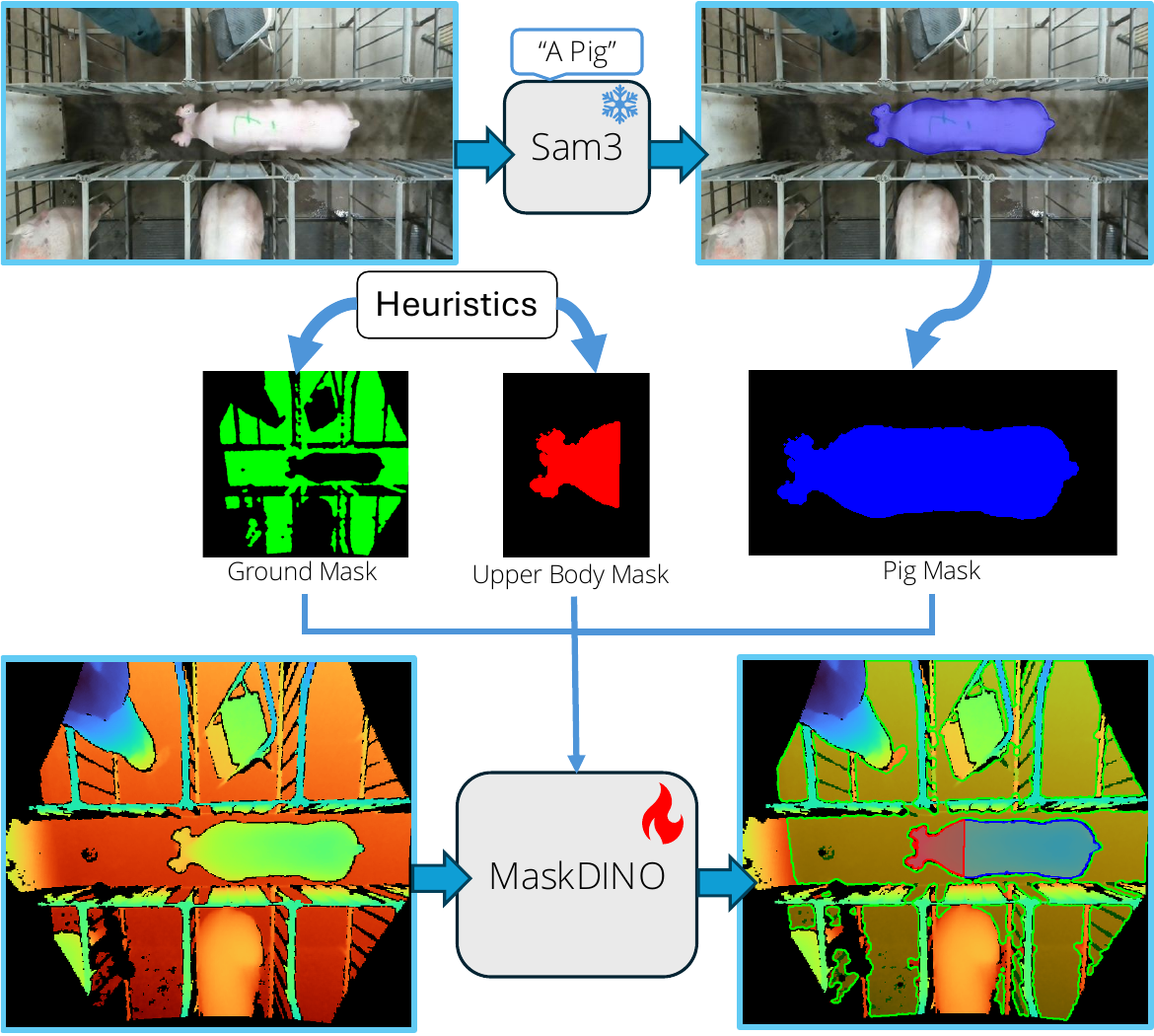}
  \caption{Segmentation pipeline overview. SAM3 generates a whole-pig mask from an RGB frame with the text prompt ``A Pig,'' which is projected to the depth camera. The ground mask is obtained via RANSAC plane fitting, and the upper-body mask is derived from the front 30\% of the pig along the heading axis. These pseudo labels train a MaskDINO model to predict all three classes from depth input alone.}
  \label{fig:segmentation_pipeline}
\end{figure}

\subsection{Segmentation}
Our segmentation pipeline produces three masks---whole pig, upper body, and ground---through a semi-automated labeling process followed by model distillation (\cref{fig:segmentation_pipeline}).

\paragraph{Whole-pig mask.} We pass each RGB frame to SAM3~\cite{Carion2025SAM3} with the text prompt ``A Pig.'' The resulting pig mask is projected from the RGB camera to the depth camera using the known camera parameters (intrinsics and extrinsics).

\paragraph{Ground mask.} We manually select three points on the ground surface and use RANSAC to fit a ground plane from these seed points.

\paragraph{Upper-body mask.} We fit a minimum-area bounding rectangle to the whole-pig mask. The short edge whose pixels have lower $z$-axis values (i.e., closer to the ground, corresponding to the head end) is identified as the front of the pig. We then take the 30\% of pig-mask pixels closest to this front edge to form the upper-body mask. This mask indicates the heading direction of the animal.

These three masks were generated in a semi-automated fashion on a small subset of the dataset to serve as pseudo ground-truth labels. We then use these labels to train a MaskDINO~\cite{li2023maskdino} model that predicts all three mask classes directly from the depth channel. Distillation to MaskDINO is necessary for two reasons: (1)~SAM3 does not perform well on raw depth images, and (2)~SAM3 cannot reliably detect the front side of the pig or segment the ground plane. By training MaskDINO on the pseudo labels derived from RGB-based SAM3 predictions, we obtain a model that operates on depth input alone at inference time, removing the need for RGB data. We exclude frames where MaskDINO fails to detect a pig or produces ambiguous masks using guardrail heuristics.

\begin{figure}[t]
  \centering
  \begin{subfigure}[t]{0.48\columnwidth}
    \centering
    \includegraphics[width=\linewidth]{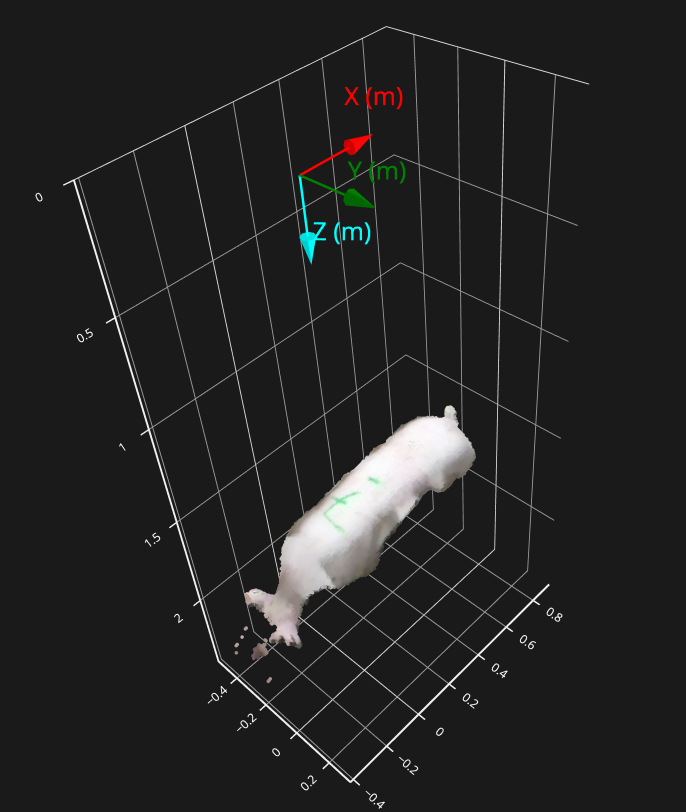}
    \caption{RGB-D point cloud}
    \label{fig:pointcloud}
  \end{subfigure}
  \hfill
  \begin{subfigure}[t]{0.48\columnwidth}
    \centering
    \includegraphics[width=\linewidth]{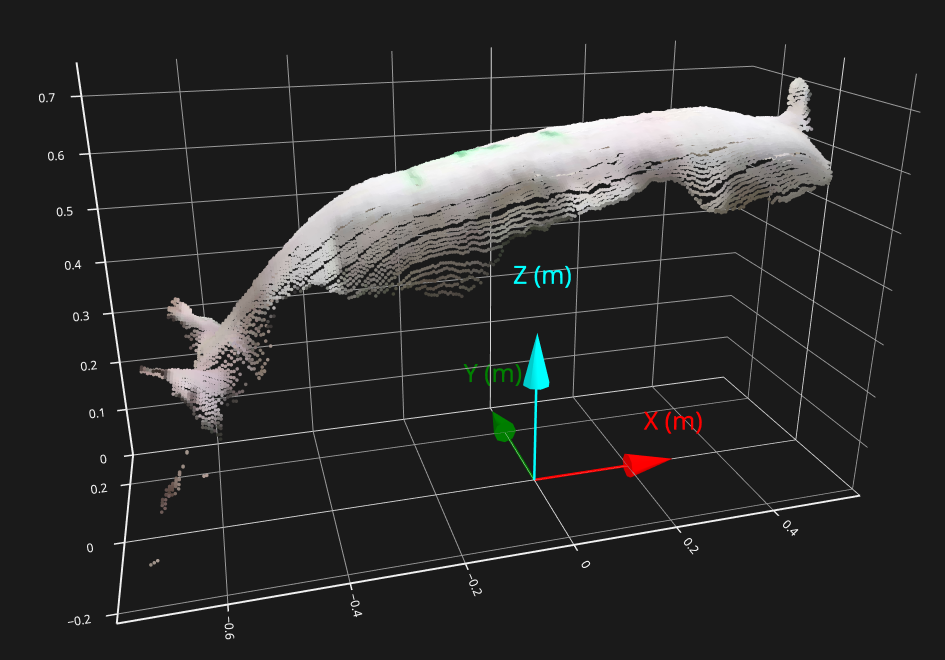}
    \caption{Height map after ground-plane removal}
    \label{fig:heightmap}
  \end{subfigure}
  \caption{From raw capture to height map. (a)~The RGB-D point cloud captured by a ceiling-mounted depth camera. (b)~The resulting height map after segmentation, ground-plane subtraction retaining only the dorsal surface profile.}
  \label{fig:pointcloud_heightmap}
\end{figure}

\subsection{Ground-plane removal and height-map generation}
We use the predicted ground mask to select reliable surface points for plane fitting. A multi-candidate RANSAC approach fits the ground plane ($z = ax + by + c$) from these points, and we subtract the fitted plane from the segmented pig region, converting absolute depth into height above ground. This normalizes for camera mounting height and floor-level variations across sessions. We then project the 3D pig points onto the ground plane and bin them into a regular grid with $1\,\text{cm} \times 1\,\text{cm}$ pixel size, where each cell stores the maximum height. The resulting height map is cropped to a standardized $96 \times 224$ resolution by centering on the pig with a dynamic margin of 64 pixels on each side of the vertical axis, retaining only the dorsal surface profile. Non-pig pixels are set to zero. We process both \blindswitch{Site 1}{MSU} and \blindswitch{Site 2}{UNL} datasets to identical dimensions to ensure compatibility for combined training.

\begin{figure}
  \centering
  \includegraphics[width=0.47\textwidth]{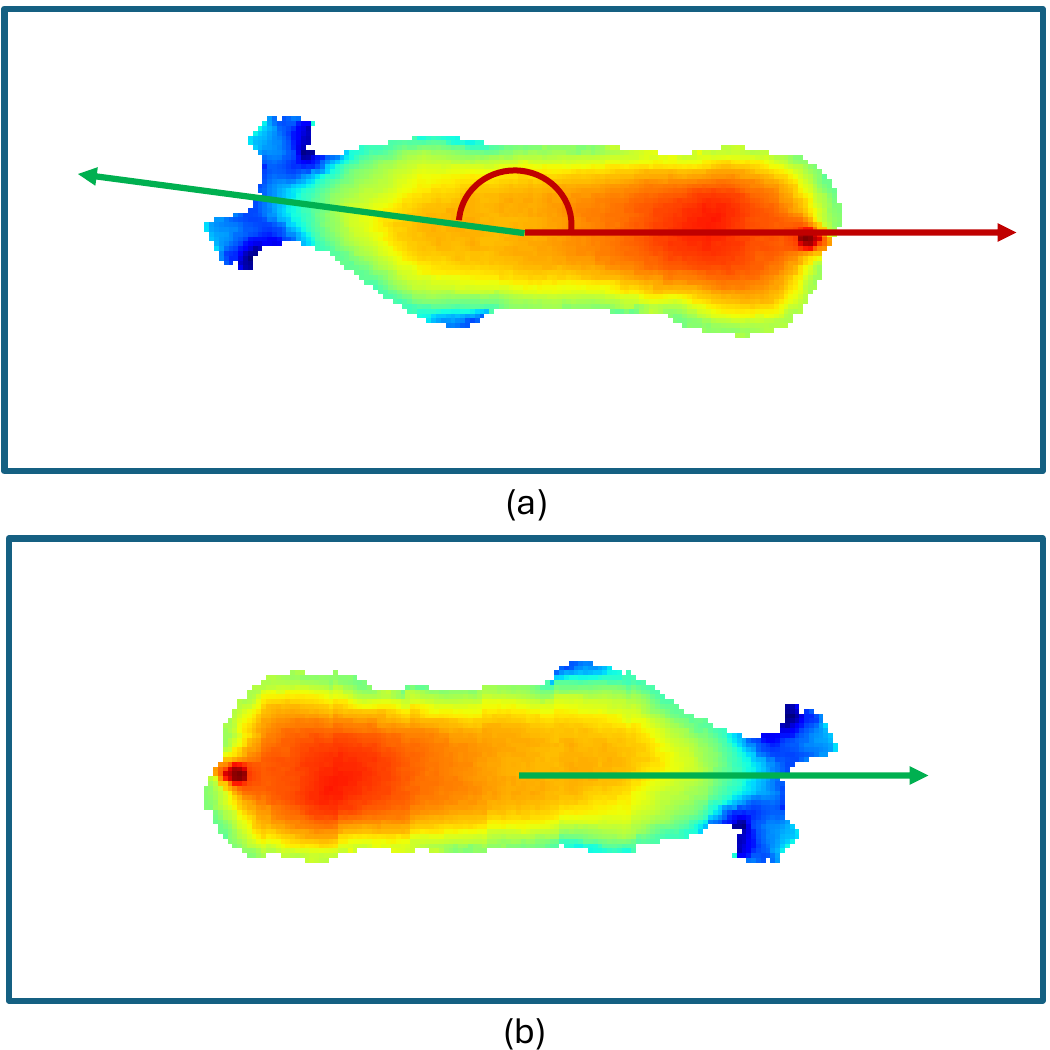}
  \caption{Heading angle estimated from upper-body mask in (a) and then used to rotate the height map in (b) so that the pig faces to the right. This eliminates pose ambiguity due to walking angle and standardizes the spatial arrangement of anatomical features across frames.}
  \label{fig:heightmap_samples}
\end{figure}
\subsection{Outlier filtering and rotation correction}
We remove frames with anomalous segmentation area (outside the interquartile range of the recording's frame areas), as these typically correspond to partial occlusions, multiple pigs in frame, or segmentation artifacts. We estimate the pig's principal body axis from the whole-pig mask via minimum-area bounding rectangle fitting, while the upper-body mask determines the heading direction. Together, these cues rotate each height map so that the pig faces to the right, eliminating pose ambiguity due to walking angle.

\subsection{Stage 1 alternatives: pruned MaskDINO and UNet}
\label{sec:supp_pruned_maskdino}

The original Stage 1 segmenter is a MaskDINO~\cite{li2023maskdino} with a ResNet-50 backbone, 300 object queries, and 9 transformer decoder layers: a configuration sized for the COCO-scale detection regime in which the architecture was designed. We chose a detection-based architecture over a pure semantic segmenter because production deployments encounter non-pig content (handlers, feeders, partial pigs at the FOV edge, and empty pens during scanning gaps). A detection-style head whose loss trains it to suppress ``no-object'' matches has stronger out-of-distribution behavior on these frames than a foreground segmenter, which is trained only to color in pig pixels. Our segmentation task is far smaller than COCO (one pig per frame, three semantic classes: pig, upper body, ground), so the original MaskDINO is over-dimensioned, and most of its 106.92\,ms / frame inference cost is paying for unused capacity.

\paragraph{Pruning rationale.} To preserve the detection-based inductive bias while removing the unused capacity, we prune the three axes most over-dimensioned for our setting:

\begin{itemize}
\item \textbf{Backbone}: ResNet-50 $\rightarrow$ ResNet-18 (24\,M $\rightarrow$ 11\,M parameters; a lower-resolution feature pyramid is still sufficient for $576\times 640$ depth input).
\item \textbf{Object queries}: 300 $\rightarrow$ 50 (we expect at most one pig per frame; the original count was tuned for COCO's 80-class, up-to-100-detections-per-image regime).
\item \textbf{Decoder layers}: 9 $\rightarrow$ 5 (the last refinement layers contributed marginally to mask AP on our task).
\end{itemize}

The pruned variant is retrained from scratch on the same data as the original: 1213 frames with hand-annotated pig and upper-body polygons. We augment this with 318 negative frames sampled from outside the labeled pig windows of the same recordings, providing ``no-pig'' supervision on handler / empty-pen frames. The pruned MaskDINO matches the original within rounding error on the held-out segmentation eval set while halving Stage 1 inference time (\cref{tab:results}, 52.73 vs.\ 106.92\,ms / frame). The downstream regression accuracy after retraining Stage 2 on its height maps is within 0.07\,mm of the original (3.94 vs.\ 3.87\,mm overall MAE).

\paragraph{UNet alternative.} As a non-detection comparison we also train a UNet with a MobileNetV3-Small encoder and a U-Net decoder, with 2 output channels (pig, upper body); the ground mask is recovered from RANSAC on the remaining depth-valid pixels. Without object queries or set-prediction loss, the UNet has neither inductive bias for instance separation nor detection-style background suppression. We therefore expected less out-of-distribution headroom for non-pig content than either MaskDINO variant. On our held-out test set, which is drawn from the same farms and capture sessions as training, this concern does not surface: the UNet matches the pruned MaskDINO within 0.01\,mm overall MAE at $16\times$ lower Stage 1 latency (6.58\,ms / frame). The UNet is therefore the option we would deploy where latency dominates; the MaskDINO variants remain preferred when field drift or operator content at the camera is the principal risk.

\paragraph{Why all three Stage 1 backends match downstream.} The original Stage 1 was sized for capacity headroom on a much larger detection task; we suspected most of that capacity was unused here. The pruned MaskDINO and UNet results are consistent with that suspicion: at the size of our segmentation training set, downstream regression accuracy is dominated by \emph{what} Stage 1 does (segmentation, heading, ground-plane removal), not the over-parameterized architecture used to do it. Retraining Stage 2 on either alternative therefore recovers the same MAE within $\pm 0.08$\,mm.

\section{Dataset}
\label{sec:supp_dataset}

The combined dataset comprises 319 instances (116 \blindswitch{Site 1}{MSU}, 203 \blindswitch{Site 2}{UNL}) totaling 6{,}705 depth frames. \blindswitch{Site 1}{MSU} pigs were scanned across 6 dates (February--December 2025), while \blindswitch{Site 2}{UNL} pigs were scanned across 11 dates (June--December 2025).

\subsection{Ground-truth labels}
We obtain ground-truth backfat depth and loin muscle depth at the last rib via B-mode ultrasound measurements taken concurrently with depth imaging sessions. Each measurement session requires trained personnel, an expensive portable ultrasound device, and physical restraint of the animal, making large-scale label acquisition costly and logistically challenging. This practical constraint is the primary reason the dataset comprises 319 instances rather than thousands. For the \blindswitch{Site 1}{MSU} dataset, we average multiple ultrasound measurements per session to produce a single fat and loin depth value per pig; the \blindswitch{Site 2}{UNL} dataset provides one measurement per pig. Some animals were scanned at multiple dates spanning 4--6 months; because body composition changes over this interval, we treat each scan date as an independent instance.

\begin{figure*}[t]
  \centering
  \begin{subfigure}[t]{\columnwidth}
    \centering
    \includegraphics[width=\linewidth]{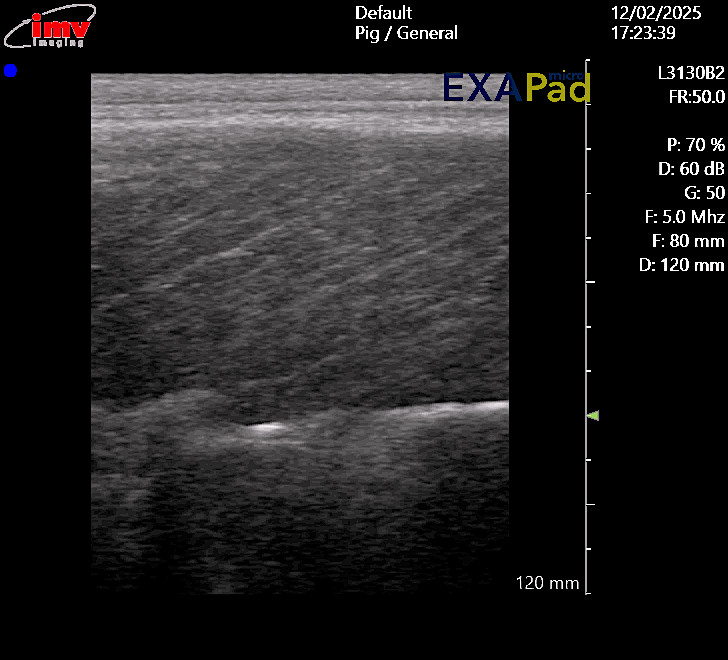}
    \caption{Raw ultrasound}
    \label{fig:raw_us}
  \end{subfigure}
  \hfill
  \begin{subfigure}[t]{\columnwidth}
    \centering
    \includegraphics[width=\linewidth]{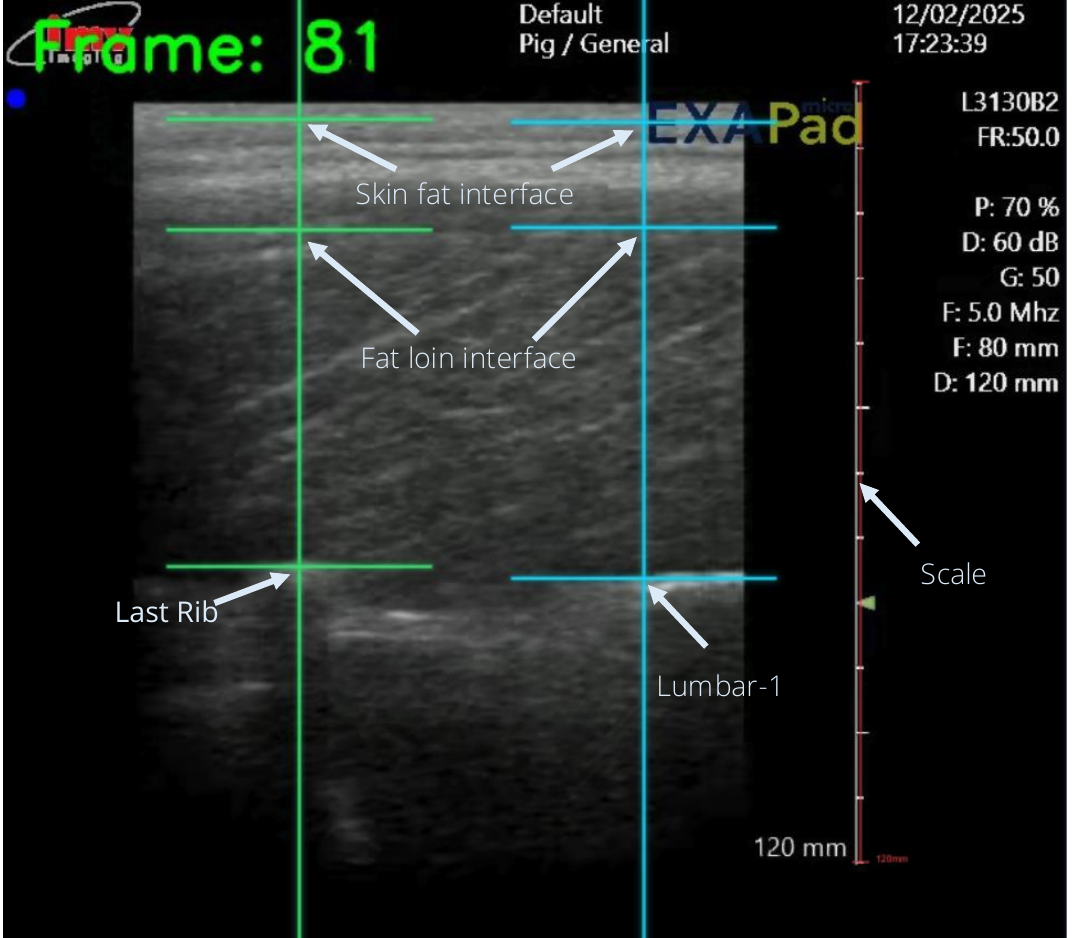}
    \caption{Annotated ultrasound}
    \label{fig:annotated_us}
  \end{subfigure}
  \caption{Ultrasound annotation process. (a) A raw ultrasound frame selected from a video where the last rib is clearly visible. (b) The corresponding annotated frame with a vertical line at the middle of the last rib and horizontal lines demarcating the skin-fat, fat-loin, and fat-rib interfaces. Annotations were performed using our custom annotation tool.}
  \label{fig:ultrasound_annotation}
\end{figure*}

For annotating the ultrasound data from \blindswitch{Site 1}{MSU}, we obtained multiple 10-second ultrasound videos per pig. An expert first reviewed the full video to select a frame where the last rib is fully and clearly visible. Within that selected frame, a vertical line was placed at the middle of the last rib, followed by three horizontal lines placed at the skin-fat interface, fat-loin interface, and fat-rib interface. If the lumbar region was visible, it was subsequently annotated as well; however, for this study, we only predict rib data with PigFormer. To streamline the expert annotation and review process, we developed a custom annotation tool; we will release this tool together with PigFormer training and evaluation code upon publication, subject to dataset-sharing restrictions from collaborating facilities. \Cref{fig:ultrasound_annotation} shows an example of a raw ultrasound frame and its corresponding annotation.

\subsection{Label statistics}

\begin{table}[h]
  \caption{Rib-site ultrasound statistics per instance (mean $\pm$ std). Total = Fat + Loin.}
  \label{tab:dataset_stats}
  \centering
  \resizebox{\columnwidth}{!}{%
  \begin{tabular}{@{}lcccc@{}}
    \toprule
    & Instances & Fat (mm) & Loin (mm) & Total (mm) \\
    \midrule
    \blindswitch{Site 1}{MSU} & 116 & $17.93 \pm 5.29$ & $54.52 \pm 6.77$ & $72.45 \pm 8.64$ \\
    \blindswitch{Site 2}{UNL} & 203 & $8.86 \pm 4.25$ & $40.94 \pm 6.91$ & $49.79 \pm 8.97$ \\
    \bottomrule
  \end{tabular}%
  }
\end{table}

The two sites exhibit distinct label distributions (\cref{tab:dataset_stats}), reflecting differences in breed, age, and feeding regimen. \blindswitch{Site 1}{MSU} pigs carry roughly twice the backfat and substantially deeper loins than \blindswitch{Site 2}{UNL} pigs. Inter-pig standard deviation is comparable across sites (fat: 5.29 vs.\ 4.25\,mm; loin: 6.77 vs.\ 6.91\,mm), confirming meaningful between-animal variation at both locations.

For \blindswitch{Site 1}{MSU}, 81 of 116 instances have multiple ultrasound readings per session, allowing us to quantify measurement repeatability. The mean intra-pig standard deviation is 1.30\,mm for fat, 2.02\,mm for loin, and 2.29\,mm for total tissue. This repeatability bound is important context for model evaluation: PigFormer's fat MAE of 2.43\,mm is less than twice the inherent variability of repeated manual ultrasound. The \blindswitch{Site 2}{UNL} dataset provides a single ultrasound reading per instance.

\subsection{Caliper score vs.\ ultrasound measurements}
\label{sec:supp_caliper}

\begin{figure*}[t]
  \centering
  \includegraphics[width=\textwidth]{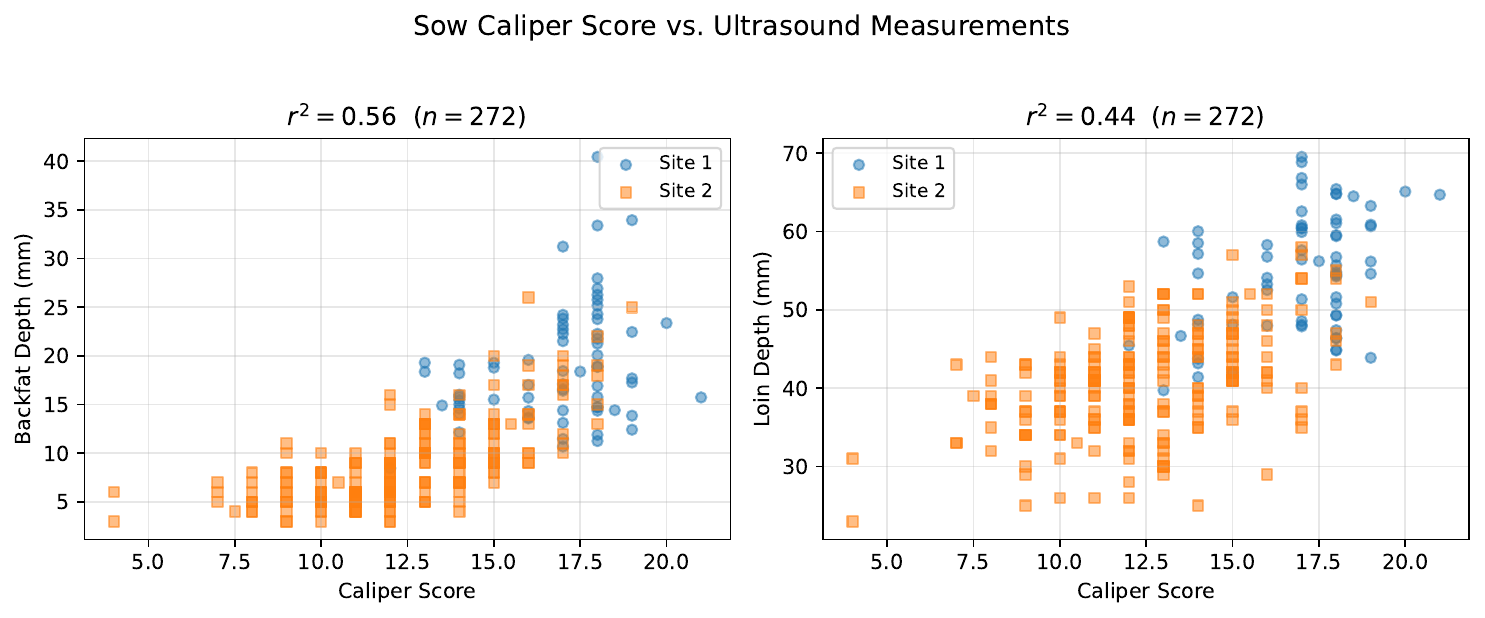}
  \caption{Sow caliper score vs.\ ultrasound-derived backfat depth and loin depth across both sites ($n=272$ instances with caliper data). Caliper scores explain only 56\% of variance in backfat ($r^2=0.56$) and 44\% in loin depth ($r^2=0.44$).}
  \label{fig:caliper_vs_ultrasound}
\end{figure*}

Sow caliper scores are widely used in production because they are fast and require no equipment, but they provide only a coarse ordinal assessment of body condition. \Cref{fig:caliper_vs_ultrasound} plots caliper score against the three ultrasound targets across both sites. The correlations are moderate at best: $r^2 = 0.56$ for backfat and $r^2 = 0.44$ for loin depth. Within individual sites the relationship is weaker still (\blindswitch{Site 1}{MSU}: $r^2 = 0.11$ for fat, $0.16$ for loin; \blindswitch{Site 2}{UNL}: $r^2 = 0.48$ for fat, $0.21$ for loin). These results confirm that caliper scores are a poor proxy for actual tissue composition, motivating our choice to regress continuous ultrasound measurements rather than predict caliper-based body condition scores.

\subsection{Sample height maps}

\section{Ablation Studies}
\label{sec:supp_ablation}

We report ablation experiments that guided the design of PigFormer. All ablations use the base PigFormer configuration (1-layer Slice Attention Encoder with dual pooling) and the combined \blindswitch{Site 1}{MSU}+\blindswitch{Site 2}{UNL} dataset (319 instances, 6{,}705 frames). Due to the broad search over hyperparameters, these supplementary ablation models were trained with a smaller training time budget than the final model reported in the main text. Evaluation uses all-frame per-instance inference with output-level averaging on a held-out test set of 79 instances.

\subsection{Slice Attention Encoder Architecture Ablations}
\label{sec:supp_arch}

To isolate the contribution of PigFormer's Slice Attention Encoder (Stage 2) from Stage 1's geometric front-end, we replace Stage 2 with two alternative encoders that share the same regression head and consume Stage 1's $96 \times 224$ height map output.

\textbf{MLP (Stage-2 alternative).} Each cross-sectional slice of the height map is passed independently through a small multi-layer perceptron to produce a per-slice feature vector. The slice-level features are concatenated, flattened, and fed to a second MLP head identical to PigFormer's to predict all three targets.

\textbf{CNN (Stage-2 alternative).} The height map is augmented with two additional channels---a binary valid-depth mask and a gradient-magnitude map---forming a three-channel image. A four-block convolutional encoder extracts a spatial feature map, which is globally pooled, projected to a 192-dimensional embedding, and passed through the same two-layer MLP regression head as PigFormer. Each block is Conv2d $\rightarrow$ BatchNorm2d $\rightarrow$ GELU, with stride 2 throughout to downsample $96 \times 224$ by $16\times$ to $6 \times 14$. Channel widths follow $3 \rightarrow 32 \rightarrow 64 \rightarrow 96 \rightarrow 128$, with a $5 \times 5$ kernel in the first block and $3 \times 3$ kernels in the remaining three. \texttt{AdaptiveAvgPool2d} collapses the $6 \times 14$ feature map to a $128$-vector, a linear layer expands it to the $192$-d embedding, and the head is Linear$(192, 256) \rightarrow$ GELU $\rightarrow$ Linear$(256, 3)$. The encoder is trained from scratch (no ImageNet pretraining); the head shape matches PigFormer's to keep capacity comparable across rows.

\Cref{tab:supp_arch} reports MAE on the held-out test set; \cref{tab:supp_arch_mape} reports the corresponding mean absolute percentage error (MAPE = MAE / mean). All three rows share Stage 1, isolating the effect of Stage 2's architecture.

\begin{table}[h]
  \caption{Slice Attention Encoder architecture ablation. All rows share Stage 1's $96 \times 224$ height map output and use the single-fold-0 / 400-epoch / input-aggregation protocol so the three encoders are directly comparable; the main-paper headline number for PigFormer (3.87\,mm) uses the stronger 4-fold ensemble / 5000-epoch / output-aggregation protocol. MAE in mm. Best results in \textbf{bold}.}
  \label{tab:supp_arch}
  \centering
  \resizebox{\columnwidth}{!}{%
  \begin{tabular}{@{}lcccc@{}}
    \toprule
    Stage 2 encoder & Fat $\downarrow$ & Loin $\downarrow$ & Total $\downarrow$ & Overall $\downarrow$ \\
    \midrule
    MLP & 2.48 & 5.35 & 4.50 & 4.11 \\
    CNN (height + geom.) & 4.62 & 5.70 & 7.72 & 6.02 \\
    \textbf{Slice Attention Encoder} (ours) & \textbf{2.35} & \textbf{4.99} & \textbf{4.39} & \textbf{3.91} \\
    \bottomrule
  \end{tabular}%
  }
\end{table}

\begin{table}[h]
  \caption{Slice Attention Encoder architecture ablation---MAPE = MAE / mean on the held-out test set. All rows share Stage 1's height map output. Best results in \textbf{bold}.}
  \label{tab:supp_arch_mape}
  \centering
  \resizebox{\columnwidth}{!}{%
  \begin{tabular}{@{}lcccc@{}}
    \toprule
    Stage 2 encoder & Fat $\downarrow$ & Loin $\downarrow$ & Total $\downarrow$ & Overall $\downarrow$ \\
    \midrule
    MLP & 14.8\% & 9.9\% & 6.4\% & 8.7\% \\
    CNN (height + geom.) & 27.6\% & 10.6\% & 11.0\% & 12.8\% \\
    \textbf{Slice Attention Encoder} (ours) & \textbf{14.1\%} & \textbf{9.3\%} & \textbf{6.2\%} & \textbf{8.3\%} \\
    \bottomrule
  \end{tabular}%
  }
\end{table}

PigFormer's Slice Attention Encoder reduces overall MAE by 5\% relative to an MLP Stage-2 alternative and 35\% relative to a CNN Stage-2 alternative. The MLP achieves competitive fat prediction (2.48\,mm) but fails on loin and total targets. Per-slice independent processing captures the localized rib-region cue for fat but loses the long-range dorsal relationships needed for the more distributed loin measurement. The CNN with auxiliary geometric channels improves over the MLP on loin and total but falls behind both alternatives on fat: local convolutions alone---without ImageNet pretraining or global attention---struggle on this 240-instance training pool. RoPE-based global attention provides the right inductive bias for height-map dorsal regression at this dataset scale.

\subsection{Classical-ML lower bound: do global statistics suffice?}
\label{sec:supp_classical}

To check whether the task requires the slice-attention encoder, or whether a simple regressor on global features could do most of the work, we fit two ridge regressors on handcrafted features extracted from the same input-aggregated height maps PigFormer consumes. Both use the same fold-0 train / val / test partition as \cref{tab:supp_arch}; the ridge regularization $\alpha$ is selected per target on fold-0 validation MAE.

\noindent\textbf{Global statistics (4 features).} Mean valid height, body volume ($\sum h_{r,c}$), body area (number of nonzero pixels), and max height. These are the most direct scalar descriptors of overall pig size and shape that a simple model could extract from the height map.

\noindent\textbf{Handcrafted geometry (15 features).} The four global statistics plus body length, body width, height std, p50 height, p97 height, length/width aspect ratio, and five along-spine zone-mean heights.

\Cref{tab:supp_classical} reports MAE on the held-out test set, with the predict-the-train-mean column included as the no-features floor.

\begin{table}[h]
  \caption{Classical-ML lower bound on the held-out test set (MAE in mm). All rows share Stage 1's input-aggregated height map and the fold-0 train / val / test partition; ridge regressors are closed-form fits with $\alpha$ selected per target on the fold-0 validation set. PigFormer row is the Slice Attention Encoder from \cref{tab:supp_arch} reproduced for context.}
  \label{tab:supp_classical}
  \centering
  \resizebox{\columnwidth}{!}{%
  \begin{tabular}{@{}lcccc@{}}
    \toprule
    Stage 2 model (\#features) & Fat $\downarrow$ & Loin $\downarrow$ & Total $\downarrow$ & Overall $\downarrow$ \\
    \midrule
    Predict train mean (0) & 4.80 & 10.05 & 13.68 & 9.51 \\
    Ridge, global statistics (4) & 4.69 & 9.81 & 12.90 & 9.13 \\
    Ridge, handcrafted geometry (15) & 2.83 & 7.53 & 8.11 & 6.16 \\
    PigFormer (ours) & \textbf{2.35} & \textbf{4.99} & \textbf{4.39} & \textbf{3.91} \\
    \bottomrule
  \end{tabular}%
  }
\end{table}

The four-feature global-statistic ridge reaches an overall MAE of 9.13\,mm: $0.38$\,mm better than predicting the training mean and $2.3\times$ worse than PigFormer. If global statistics sufficed, this row would already be close to PigFormer. It is not. This contradicts the hypothesis that the model relies on average height or body size. It also matches the Normalized Column Importance analysis in \cref{sec:supp_attention}, which shows the encoder localizes to specific spine regions rather than computing a global average.

Extending the feature set to 15 hand-engineered geometric descriptors (per-zone means, length, width, percentiles, aspect) reduces overall MAE to 6.16\,mm, comparable to the CNN Stage-2 alternative in \cref{tab:supp_arch} (6.02\,mm). Both still fall short of PigFormer by $\geq 2.2$\,mm, with the largest gap on loin and total: the two targets that depend on regional anatomy rather than bulk. The single-layer slice-attention encoder therefore contributes a geometric inductive bias on top of Stage 1 that neither handcrafted features nor convolutional layers capture at this dataset scale.

\subsection{Frame sampling strategy}
\Cref{tab:abl_frames} shows the effect of the number of frames averaged at the pixel level per instance during training. At inference, all frames are used with output-level averaging.

\begin{table}[h]
  \caption{Effect of frame sampling strategy on MAE (mm). Single random frame per epoch serves as the base configuration.}
  \label{tab:abl_frames}
  \centering
  \resizebox{\columnwidth}{!}{%
  \begin{tabular}{@{}lcccc@{}}
    \toprule
    Frames/Epoch & Fat $\downarrow$ & Loin $\downarrow$ & Total $\downarrow$ & Overall $\downarrow$ \\
    \midrule
    1 (random) & \textbf{2.38} & 5.34 & 4.29 & 4.00 \\
    3 & 2.58 & \textbf{4.81} & \textbf{3.99} & \textbf{3.79} \\
    5 & 2.62 & 5.51 & 4.24 & 4.12 \\
    8 & 2.65 & 5.84 & 5.16 & 4.55 \\
    All & 3.69 & 6.88 & 5.99 & 5.52 \\
    \bottomrule
  \end{tabular}%
  }
\end{table}

Averaging 3 frames achieves the best overall MAE (3.79\,mm). Performance degrades monotonically beyond 3 frames. Using all frames is worst (5.52\,mm), confirming that pixel-averaging many frames creates deformed height maps. Single random frame per epoch (4.00\,mm) remains a strong baseline due to implicit data augmentation across epochs.

\subsection{Data augmentation}
To understand which augmentations contribute, \Cref{tab:abl_aug_individual} tests each one individually and in combination using full 4-fold cross-validation with ensemble evaluation.

\begin{table}[h]
  \caption{Individual augmentation effects (full 4-fold CV with smaller training time budget).}
  \label{tab:abl_aug_individual}
  \centering
  \resizebox{\columnwidth}{!}{%
  \begin{tabular}{@{}lccccc@{}}
    \toprule
    Augmentation & Fat $\downarrow$ & Loin $\downarrow$ & Total $\downarrow$ & Overall $\downarrow$\\
    \midrule
    None (baseline) & \textbf{2.45} & 5.58 & 4.63 & 4.22 \\
    Vertical flip & 2.52 & 5.35 & \textbf{4.34} & \textbf{4.07} \\
    Horizontal flip & 2.47 & 6.41 & 5.70 & 4.86 \\
    Translation ($\pm$10px) & 2.58 & 5.61 & 4.92 & 4.37 \\
    Rotation ($\pm$5$^\circ$) & 2.46 & 5.56 & 4.62 & 4.21 \\
    VFlip + Trans + Rot & 2.64 & \textbf{5.22} & 4.39 & 4.08\\
    All four & 2.66 & 5.77 & 5.00 & 4.48 \\
    \bottomrule
  \end{tabular}%
  }
\end{table}

Vertical flip is the most beneficial augmentation (4.07\,mm vs.\ 4.22\,mm baseline), achieving the best overall MAE among all configurations tested. Horizontal flip hurts significantly (4.86\,mm)---likely because head-to-tail orientation encodes meaningful spatial information for body composition prediction. Rotation provides minimal benefit, and translation slightly hurts. Combining vertical flip with translation and rotation yields competitive performance (4.08\,mm), but adding horizontal flip degrades performance even in combination (4.48\,mm). Fat MAE is robust across augmentations ($\sim$2.45--2.66\,mm); the differences are driven primarily by loin and total predictions.

\subsection{Training data composition}
\Cref{tab:abl_site} compares site-specific and combined training. All models are evaluated on the same mixed-site test set (26 \blindswitch{Site 1}{MSU} + 53 \blindswitch{Site 2}{UNL} instances).

\begin{table}[h]
  \caption{Effect of training data composition on MAE (mm).}
  \label{tab:abl_site}
  \centering
  \resizebox{\columnwidth}{!}{%
  \begin{tabular}{@{}lccccc@{}}
    \toprule
    Training Data & \#Train & Fat $\downarrow$ & Loin $\downarrow$ & Total $\downarrow$ & Overall $\downarrow$ \\
    \midrule
    \blindswitch{Site 1}{MSU} only & 90 & 4.90 & 7.88 & 10.37 & 7.72 \\
    \blindswitch{Site 2}{UNL} only & 150 & 3.35 & 8.86 & 10.67 & 7.63 \\
    Combined & 240 & \textbf{2.38} & \textbf{5.34} & \textbf{4.29} & \textbf{4.00} \\
    \bottomrule
  \end{tabular}%
  }
\end{table}

Combined training reduces overall MAE by nearly half compared to single-site training. Neither site alone generalizes to the mixed test set, despite comparable dataset sizes. The two sites house different breeds with markedly different body compositions. \blindswitch{Site 1}{MSU} animals carry roughly twice the backfat ($17.93$ vs.\ $8.86$\,mm) and substantially greater loin depth ($54.52$ vs.\ $40.94$\,mm) than \blindswitch{Site 2}{UNL} animals. A model trained on only one site learns a narrow range of fat and loin values and fails to extrapolate to the other site's distribution. Combining both sites exposes the model to the full spectrum of body compositions encountered across breeds, ages, and management regimes, which is the primary driver of generalization rather than dataset size alone.

\subsection{Model depth}
\Cref{tab:abl_depth} shows the effect of Slice Attention Encoder depth.

\begin{table}[h]
  \caption{Effect of encoder depth on MAE (mm).}
  \label{tab:abl_depth}
  \centering
  \resizebox{\columnwidth}{!}{
  \begin{tabular}{@{}lcccc@{}}
    \toprule
    Layers & Fat $\downarrow$ & Loin $\downarrow$ & Total $\downarrow$ & Overall $\downarrow$ \\
    \midrule
    1 & 2.38 & \textbf{5.34} & \textbf{4.29} & \textbf{4.00} \\
    2 & \textbf{2.37} & 6.31 & 5.49 & 4.72 \\
    3 & 2.42 & 6.53 & 5.80 & 4.92 \\
    \bottomrule
  \end{tabular}
  }
\end{table}

A single-layer transformer achieves the best performance. Adding depth hurts loin and total predictions while fat prediction remains stable ($\sim$2.4\,mm), suggesting that fat depends on simpler spatial features. The small dataset (240 training instances) does not support deeper models.

\begin{table*}[htbp]
  \caption{Computational cost for the systems in \cref{tab:results}. Params (M) and FLOPs (G) are total system; per-frame inference time is split into Stage 1 + Stage 2 with the total in parentheses.}
  \label{tab:supp_efficiency}
  \centering
  \resizebox{\textwidth}{!}{%
  \begin{tabular}{@{}lrrrrr@{}}
    \toprule
    Method & Params (M) & FLOPs (G) & Stage 1 (ms) & Stage 2 (ms) & Total (ms) \\
    \midrule
    ViT-small (single-stage) & 21.47 & 4.21 & --- & 4.98 & 4.98 \\
    ResNet-18 (single-stage) & 11.17 & 12.78 & --- & 2.88 & 2.88 \\
    \midrule
    \textbf{PigFormer} w. MaskDINO & 52.39 & 122.28 & 106.92 & 0.50 & 107.42 \\
    \textbf{PigFormer} w. Pruned MaskDINO & 29.63 & 91.83 & 52.73 & 0.50 & 53.23 \\
    \textbf{PigFormer} w. UNet & \textbf{2.20} & \textbf{4.06} & \textbf{6.58} & \textbf{0.50} & \textbf{7.08} \\
    \bottomrule
  \end{tabular}%
  }
\end{table*}

\subsection{Feature-level vs.\ pixel-level aggregation}
\Cref{tab:abl_agg} compares multi-frame aggregation strategies using 5 frames per sample for input.

\begin{table}[h]
  \caption{Multi-frame aggregation strategy comparison (mm), using 5 frames.}
  \label{tab:abl_agg}
  \centering
  \resizebox{\columnwidth}{!}{%
  \begin{tabular}{@{}lcccc@{}}
    \toprule
    Strategy & Fat $\downarrow$ & Loin $\downarrow$ & Total $\downarrow$ & Overall $\downarrow$ \\
    \midrule
    Pixel-level & 2.62 & \textbf{5.51} & \textbf{4.24} & \textbf{4.12} \\
    Feature-level & \textbf{2.31} & 6.20 & 5.51 & 4.67 \\
    \bottomrule
  \end{tabular}
  }
\end{table}

Pixel-level aggregation (averaging raw frames) outperforms feature-level aggregation (averaging encoder outputs) overall. Feature-level achieves slightly better fat prediction (2.31\,mm) but loses spatial coherence important for loin and total depth. Combined with the frame sampling ablation, the optimal strategy is 3-frame pixel averaging (3.79\,mm overall).

\subsection{Computational efficiency}
\label{sec:supp_efficiency}

\Cref{tab:supp_efficiency} pairs the accuracy numbers in \cref{tab:results} with parameter count, FLOPs, and per-frame inference time. Parameters and FLOPs are reported for the full inference path (Stage 1 $+$ Stage 2 for the PigFormer rows). FLOPs are counted with \texttt{fvcore.nn.FlopCountAnalysis} at the native input resolution: $576 \times 640$ for ResNet-18, the UNet, and both MaskDINO variants; $224 \times 224$ for ViT-small (which is bilinear-resized inside its forward pass). MaskDINO totals are slight underestimates ($\sim$5--10\%) because the multi-scale deformable-attention CUDA op is not traceable by \texttt{fvcore}. Per-frame inference time is measured on A100 with batch $=$ 1 (MaskDINO Stage 1 in fp16; UNet Stage 1, the single-stage backbones, and PigFormer Stage 2 in fp32).

Three observations:

\noindent\textbf{Stage 2 is essentially free.} The Slice Attention Encoder is 0.24\,M parameters and 0.04\,GFLOPs on its own (the values above are Stage 1 $+$ Stage 2 combined); its 0.50\,ms / frame inference cost is negligible relative to Stage 1 across every variant. PigFormer's compute budget is almost entirely Stage 1.

\noindent\textbf{Pruning the segmenter is effective.} Replacing the original MaskDINO (R50, 300q, 9L) with the pruned variant (R18, 50q, 5L) cuts parameters by 43\% (52.4\,M $\rightarrow$ 29.6\,M), FLOPs by 25\% (122 $\rightarrow$ 92\,G), and Stage 1 latency by 50\% (106.92 $\rightarrow$ 52.73\,ms), all at $+0.07$\,mm overall MAE.

\noindent\textbf{The UNet variant is the smallest deployable system in the table.} At 2.20\,M parameters and 4.06\,GFLOPs (total system), it is smaller than either single-stage baseline, while delivering $3.95$\,mm overall MAE---$+0.08$\,mm relative to the original. End-to-end latency drops to $\approx 7$\,ms / frame, an order of magnitude faster than PigFormer with the original MaskDINO.

\begin{figure*}[htbp]
  \centering
  \includegraphics[width=\textwidth]{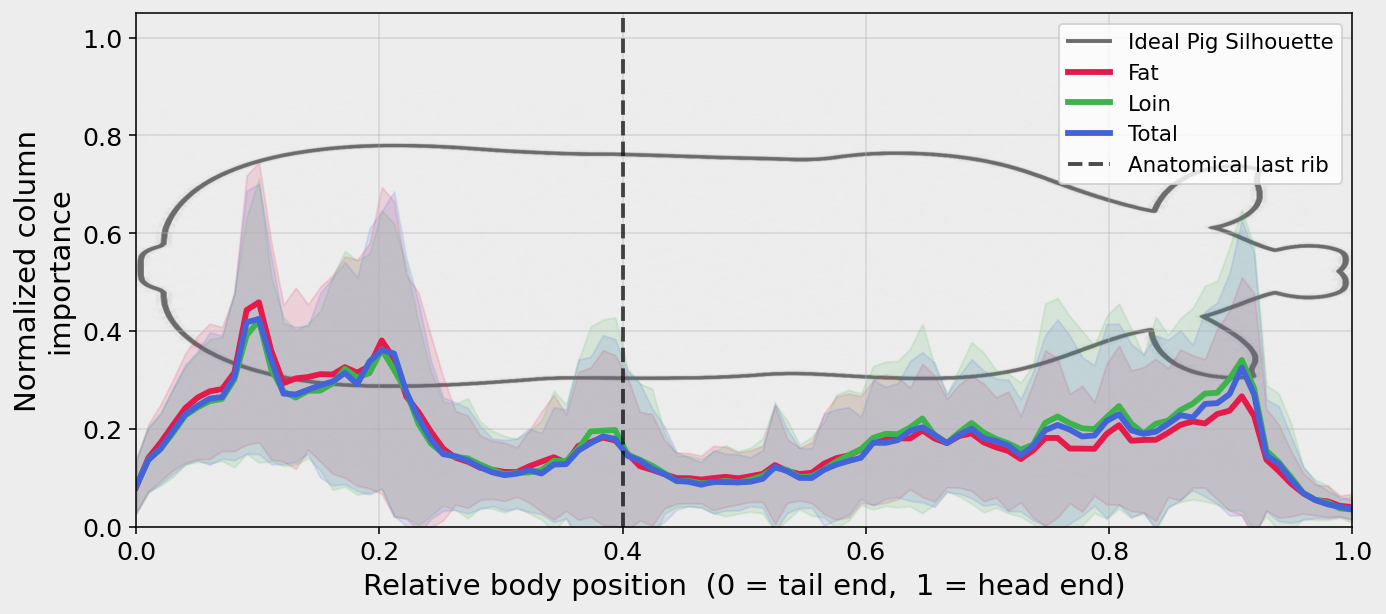}
  \caption{Population Normalized Column Importance (NCI) curve averaged over all 79 test bags with $\pm 1\sigma$ shading per target. The faded outline is the average pig silhouette in the same relative-body coordinates ($0 =$ tail, $1 =$ head). The non-uniform shape (rump peak near relative position $0.1$; rib-shoulder plateau from $0.5$ to $0.9$) rules out global-statistic regression. The anatomical last rib (dashed vertical at $0.4$) sits on the plateau's rising edge.}
  \label{fig:nci_aggregate}
\end{figure*}

\begin{figure*}[t]
  \centering
  \includegraphics[width=\textwidth]{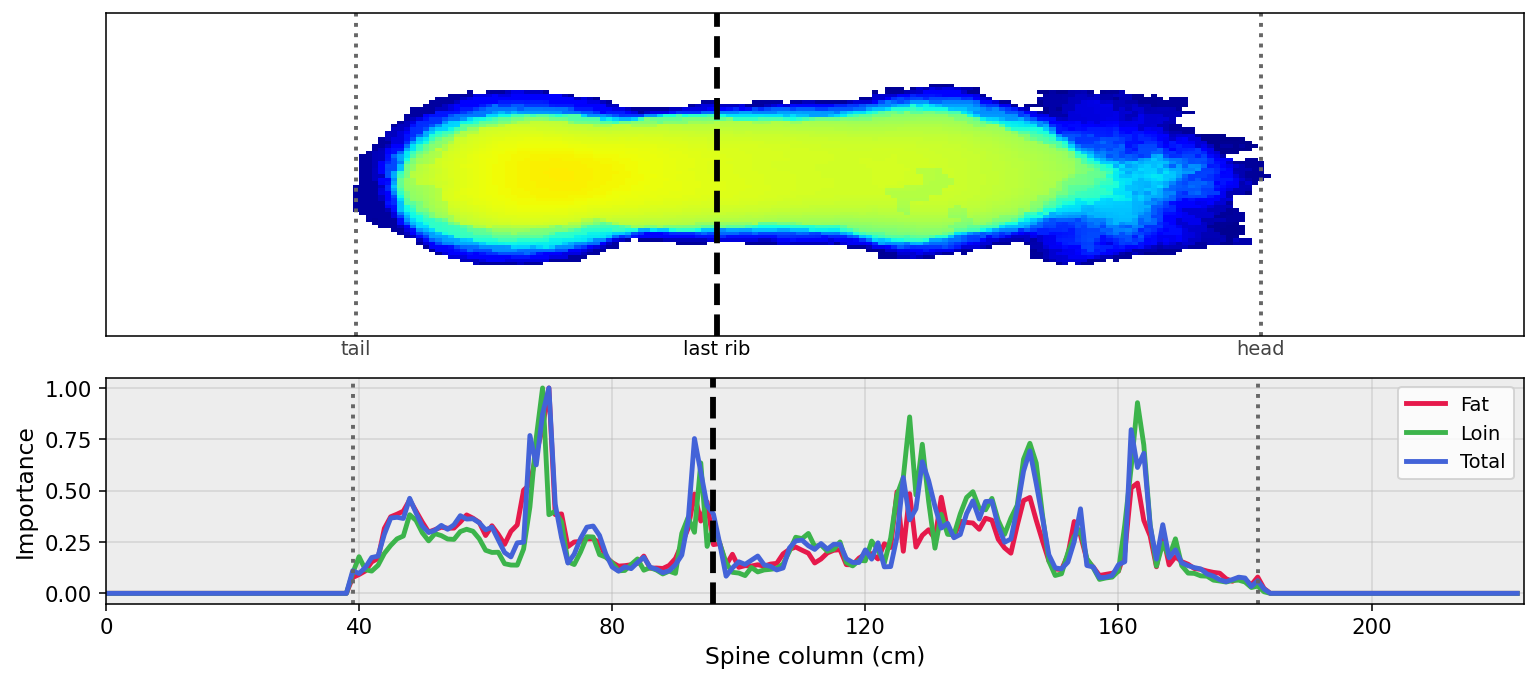}
  \caption{Per-spine-column importance for one representative test pig (input-aggregated height map). Top: height map with tail / last-rib / head columns marked. Bottom: per-target column importance. The loin curve (green) shows a localized peak near the last-rib column even when fat (red) and total (blue) are broader, consistent with back-muscle depth being a more spatially concentrated signal than fat thickness.}
  \label{fig:nci_single}
\end{figure*}

\section{Does the encoder localize to pig anatomy?}
\label{sec:supp_attention}

The Slice Attention Encoder has a single transformer layer, so one can ask whether it has learned anatomical structure or is regressing global statistics of the height map (mean height, body volume, body area). Under the global-statistic hypothesis, an input-attribution map should be approximately uniform across the pig. We show below that the population-averaged attribution is non-uniform and concentrates at anatomically meaningful positions.

\paragraph{Normalized Column Importance (NCI).} For an input height map $x\in\mathbb{R}^{96\times 224}$ (rows index the lateral axis, columns index the spine axis) and target $i\in\{\mathrm{fat},\mathrm{loin},\mathrm{total}\}$, we compute a SmoothGrad-$\times$-Input attribution:
\begin{align}
\bar g_i(x) &= \frac{1}{K}\sum_{k=1}^{K}\Bigl|\nabla_x\,f_i\!\bigl(x + \sigma\,\varepsilon_k\odot M(x)\bigr)\Bigr|, \\
A_i(x) &= |x|\odot\bar g_i(x),
\end{align}
where $f_i$ is the $i$-th model output, $\varepsilon_k\sim\mathcal{N}(0,I)$, $M(x)=\mathbf{1}(x\neq 0)$ restricts the SmoothGrad noise to pig pixels, $K=20$, and $\sigma=0.04$\,m (4\,cm of height noise). Multiplying by $|x|$ zeros out background pixels: the attribution then measures what the model uses in the \emph{current} input rather than counterfactual sensitivity to non-existent depth. This distinction matters at the body boundary, where vanilla SmoothGrad otherwise dominates because that is where adding depth most affects the prediction.

The per-spine-column importance is the cross-sectional sum, normalized by its peak:

\begin{align}
C_i[c]=\sum_{r=0}^{95}A_i(x)[r,c] \\
\mathrm{NCI}_i[c]=\frac{C_i[c]}{\max_{c'}C_i[c']}\in[0,1].
\end{align}

\paragraph{Aggregation across the test set.} For each of the 79 test bags $b$ we (i)~average all bag frames into the input-aggregated height map $x_b$ (paper test protocol), (ii)~locate the pig spine extent $[\mathrm{tail}_b,\mathrm{head}_b]$ as the leftmost and rightmost columns whose vertical pig mass exceeds 5\% of the peak (pigs face right after rotation, so the rightmost pig column is the head tip), and (iii)~resample $\mathrm{NCI}_i$ on $[\mathrm{tail}_b,\mathrm{head}_b]$ onto a fixed 100-bin grid in relative-body coordinates ($0 =$ tail end, $1 =$ head end). The population curve in \cref{fig:nci_aggregate} averages across all bags with $\pm 1\sigma$ shading.

\paragraph{Findings.} The aggregate NCI curve is non-uniform, with multiple peaks separated by valleys, which rules out the global-statistic hypothesis. The two most prominent regions are a sharp tail/rump peak near relative position $0.1$ and a broad rib-shoulder plateau spanning approximately $0.5$--$0.9$. The anatomical last rib (relative position $0.4$, computed as 60\% of body length from the head toward the tail) sits on the rising edge of this plateau. All three targets share the same overall shape, consistent with (a)~the encoder's shared mean+max pool feeding all three regression heads, and (b)~the strong target correlations (total $=$ fat $+$ loin by construction).

The single-pig example in \cref{fig:nci_single} shows the same structure resolved on one test instance. The loin curve spikes locally near the last-rib column while fat and total are broader: back-muscle depth is a more spatially localized signal than fat thickness. This is consistent with the higher per-target MAE on loin reflecting genuine geometric difficulty rather than the model averaging globally.